\documentclass[pdflatex,a4paper,fleqn]{cas-dc}

\usepackage[numbers]{natbib}
\usepackage{amsmath,amssymb,amsfonts}
\usepackage{pifont}
\newcommand{\xmark}{\ding{55}}
\newcommand{\cmark}{\checkmark}
\usepackage{xcolor}
\usepackage{placeins}
\usepackage{stfloats} 
\usepackage{caption}
\usepackage{cuted}

\newcommand{\added}[1]{#1}

\begin{document}
\let\WriteBookmarks\relax
\def\floatpagepagefraction{1}
\def\textpagefraction{.001}

\shorttitle{}    


\title [mode = title]{TCSA-UDA: Text-Driven Cross-Semantic Alignment for Unsupervised Domain Adaptation in Medical Image Segmentation}  





\author[1]{Lalit Maurya}[orcid=0000-0003-2234-8692]

\cormark[1]


\ead{lalit.maurya@port.ac.uk}




\author[1]{Honghai Liu}[orcid=0000-0002-2880-4698]


\ead{honghai.liu@port.ac.uk}



\affiliation[1]{organization={School of Computing, University of Portsmouth},
            city={Portsmouth},
            postcode={PO1 3HE}, 
            country={UK}}
\author[2]{Reyer Zwiggelaar}[orcid=0000-0002-4360-0896]


\ead{rrz@aber.ac.uk}



\affiliation[2]{organization={Department of Computer Science, Aberystwyth University},
            city={Aberystwyth},
            postcode={SY23 3DB}, 
            country={UK}}
\cortext[1]{Corresponding author}



\begin{abstract}
Unsupervised domain adaptation (UDA) for medical image segmentation remains challenging due to substantial domain shifts across imaging modalities, such as CT and MRI. Although recent vision-language representation learning methods have shown promise in medical image analysis, their role in cross-modality UDA segmentation remains underexplored. To address this problem, we propose TCSA-UDA, a Text-driven Cross-Semantic Alignment framework that uses modality-aware textual prompting to guide domain-invariant visual representation learning. Specifically, we introduce a vision-language covariance cosine loss (VLCoL) that aligns inter-class visual feature relationships with text-derived semantic relationships, encouraging the image encoder to learn semantically structured and modality-robust representations. In addition, we incorporate a prototype alignment module to reduce residual class-level discrepancies between source and target domains by aligning high-level class prototypes. Extensive experiments on cross-modality cardiac, abdominal, and brain tumor segmentation benchmarks demonstrate that TCSA-UDA consistently improves adaptation performance and outperforms state-of-the-art UDA methods. These results highlight the potential of language-driven semantic guidance for domain-adaptive medical image segmentation.
The code is available at \url{https://github.com/lalitmaurya47/TCSA_UDA}
\end{abstract}





\maketitle

\section{Introduction}
Medical image segmentation employs modalities like CT and MRI to offer detailed anatomical insights for diagnosis and surgical planning. While CT provides high spatial resolution and bone detail, MRI excels in soft tissue contrast \cite{florkow2022magnetic, jacobson2026prostate}. Although clinicians can intuitively integrate multi-modal data, models trained on a single modality often fail to generalize across domains due to differences in imaging protocols, scanners, and institutions. Fully-supervised deep learning methods perform well in-domain but are sensitive to domain shifts, which are common in real-world medical data. Annotating data from each new domain is labor-intensive and often impractical in clinical workflows. Moreover, domain variability in medical imaging, stemming from acquisition settings and equipment differences, poses a significant challenge to model robustness \cite{10604846}. Thus, there is a critical need for domain-adaptive models capable of generalizing across diverse data distributions without relying on extensive re-annotation, ensuring reliable and scalable deployment in clinical environments. 
\begin{figure*}
    \centering
    \includegraphics[width=0.7\linewidth]{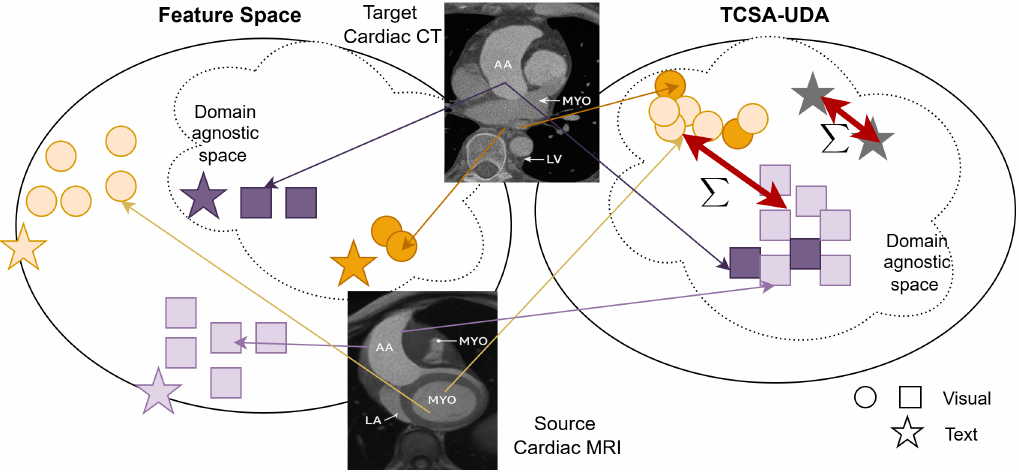}
    \caption{Class-specific visual features (squares and circles) from source and target domains are aligned with shared semantic class embeddings derived from modality-aware text prompts (stars) via covariance matching. Red bidirectional arrows represent cross-modal covariance learning, encouraging semantic consistency across domains within a shared feature space while preserving modality context.}
    \label{fig_embed}
\end{figure*}
Recently, Unsupervised Domain Adaptation (UDA) in computer vision has gained momentum due to its ability to transfer knowledge from labeled source domains to unlabeled target domains, effectively reducing reliance on manual annotation \cite{chen2019synergistic,liu2022deep}. In particular, image-level approaches focus on appearance adaptation using image-to-image translation \cite{8237506}, while feature-level methods aim to learn domain-invariant representations \cite{Feng_Ju_Wang_Song_Zhao_Ge_2023}.  Many UDA methods employ adversarial training or distance-based metrics like maximum mean discrepancy (MMD) \cite{long2015learning} to align distributions across domains \cite{8988158,zhao2022uda}. Some strategies also perform adaptation in the output space to align the spatial structure of predictions \cite{sun2022attention,Feng_Ju_Wang_Song_Zhao_Ge_2023}. However, these often rely on cross-entropy loss, which may lead to overconfident and miscalibrated predictions in the target domain, particularly due to the absence of target supervision.  Self-training under a teacher–student framework has been explored as an alternative, where pseudo-labels from a teacher model guide the student model. Yet, self-training is highly sensitive to label noise, and standard data augmentation techniques often fail to capture the complex variations in medical imaging domains \cite{Zou_2018_ECCV,Li_2019_CVPR}.

Critically, current UDA pipelines lack robust semantic consistency across domains, particularly when adapting across heterogeneous imaging modalities. We posit that semantic guidance grounded in shared anatomical concepts, rather than purely visual alignment, is a missing element. Unlike visual features, textual class semantics (e.g., “left atrium”, “tumor”) are inherently domain-invariant, as they describe anatomical entities that remain consistent across modalities. However, the expression of these semantics can be modality-aware, reflecting differences between imaging types such as CT and MRI. 


Motivated by the domain-invariant nature of anatomical semantics, we propose incorporating text embeddings to guide visual representation learning through text-driven cross-semantic alignment with adversarial training (Fig. \ref{fig_embed}). Modality-aware prompts are used to encode both anatomical class information and imaging-modality context, while visual-language alignment encourages the segmentation network to learn semantically meaningful and modality-robust feature representations. In particular, we introduce a vision-language covariance cosine loss (VLCoL) to align inter-class visual feature relationships with text-derived semantic relationships. This helps preserve anatomical semantic structure across imaging modalities and improves cross-domain generalization in the absence of target-domain annotations.

The main contributions of this work are summarized as follows:

\begin{itemize}

\item We propose TCSA-UDA, a text-driven cross-semantic adaptation framework for unsupervised cross-modality medical image segmentation.
\item We introduce VLCoL, a vision-language covariance cosine loss that aligns visual inter-class feature relationships with text-derived anatomical semantic relationships to improve domain-invariant representation learning. 
\item We integrate modality-aware textual prompting, Transformer based visual language fusion, and dynamic convolution to inject modality-aware anatomical semantics into the segmentation network. Prototype alignment is used as a complementary class-level adaptation module to reduce residual source-target discrepancies. 
\item We validate the proposed framework on cardiac, abdominal organ, and brain tumor cross-modality segmentation tasks and conduct extensive ablation studies to analyze the contribution of the key components.

\end{itemize}

\section{Related Work}

\subsection{Unsupervised Domain Adaptation}
UDA techniques aim to minimize performance discrepancies between source and target domains caused by domain shifts. These techniques can be categorized into image-level adaptation, feature-level adaptation, output-level adaptation, and source-free domain adaptation. Image-level methods such as CycleGAN \cite{8237506} transform source images to mimic the target domain, improving segmentation training. Feature-level approaches utilize adversarial learning to extract domain-invariant features \cite{long2015learning,zhang2020collaborative}. Output-level strategies focus on aligning spatial layouts and semantic structures \cite{tsai2018learning,vu2019advent}. Recent advancements incorporate more adaptive strategies. SECASA \cite{Feng_Ju_Wang_Song_Zhao_Ge_2023} introduces entropy-guided constraints and semantic alignment, while DPCL  \cite{en2024unsupervised} leverages evolving class prototypes and contrastive learning to enhance discriminative representation. Additionally, Transformer-based architectures such as DAFormer \cite{9879466} and MA-UDA \cite{10273225} demonstrate improved generalization by integrating hierarchical attention mechanisms and meta-adaptation for cross-modality segmentation. Beyond conventional UDA settings, source-free domain adaptation methods have recently attracted attention because they adapt models to target domains without requiring access to source-domain data. In this category, SFDA-CCRC \cite{ma2025source} introduces contrastive class-relationship consistency for source-free cross-modality cardiac segmentation, while IPLC \cite{zhang2024iplc} employs a SAM-guided iterative pseudo-label correction framework to improve target-domain adaptation in medical image segmentation. IPLC+ \cite{zhang2025iplc+} further extends IPLC by enhancing target-domain pseudo-label quality through a reliable SAM-based pseudo-label generation strategy. Deng et al. \cite{10604846} proposed an unsupervised electron microscopy (EM) image denoising method using domain alignment and adversarial training to disentangle content and noise.
However, most existing approaches are limited to visual modalities and fail to utilize semantic cues embedded in language. The integration of multimodal information, particularly from vision-language models, presents a promising but underexplored direction for enhancing domain adaptation in medical image segmentation.

\subsection{Text-Assisted Medical Image Segmentation}
Recent advances in medical image segmentation have demonstrated the effectiveness of integrating textual and visual information to boost performance. TGANet \cite{tomar2022tganet} leveraged text embeddings for colonoscopy segmentation, surpassing image-only methods. Zhong et al. \cite{zhong2023ariadne} showed that language-guided segmentation improves Dice scores while reducing data demands. Liu et al. \cite{liu2023m} introduced frozen language models to stabilize training and enrich latent representations. Chen et al. \cite{chen2023generative} extended vision-language pretraining to 3D imaging using synthetic text annotations to address data scarcity. \added{In natural image segmentation, CLIPSeg \cite{luddecke2022image} demonstrated that CLIP embeddings can guide open-vocabulary segmentation without domain-specific training. MaskCLIP \cite{zhou2022extract} extended this to dense prediction tasks, showing that CLIP's patch-level features carry rich semantic information transferable across domains. For domain adaptation specifically, PADCLIP \cite{Lai_2023_ICCV} leverages CLIP's text encoder to generate domain-invariant textual prompts at inference time, reducing the visual domain gap in natural image UDA. DAD-CLIP \cite{Singha_2023_ICCV} further adapts CLIP in prompt space by learning domain-invariant and class-generalizable prompts conditioned on image style and content, demonstrating the potential of prompt-based adaptation for addressing domain shift. In the medical domain, MedCLIP \cite{wang2022medclip} pre-trains a vision-language model on medical image-text pairs, demonstrating transferable representations across imaging modalities. UniverSeg \cite{Butoi_2023_ICCV} showed that cross-modal segmentation can be guided by example-based conditioning without domain-specific retraining. }

However, these methods are primarily designed for open-vocabulary segmentation, prompt adaptation, or supervised / single-domain medical segmentation, and they do not explicitly use language-derived inter-class relations as a training-time semantic prior for unsupervised cross-modality segmentation. To address this, our method incorporates a vision–language covariance cosine loss to directly align image encoder features with inter-class textual semantics, fostering modality-invariant and semantically meaningful feature representations that enhance cross-domain generalization in medical segmentation tasks.

\begin{figure*}[ht!]
    \centering
    \includegraphics[width=1\linewidth]{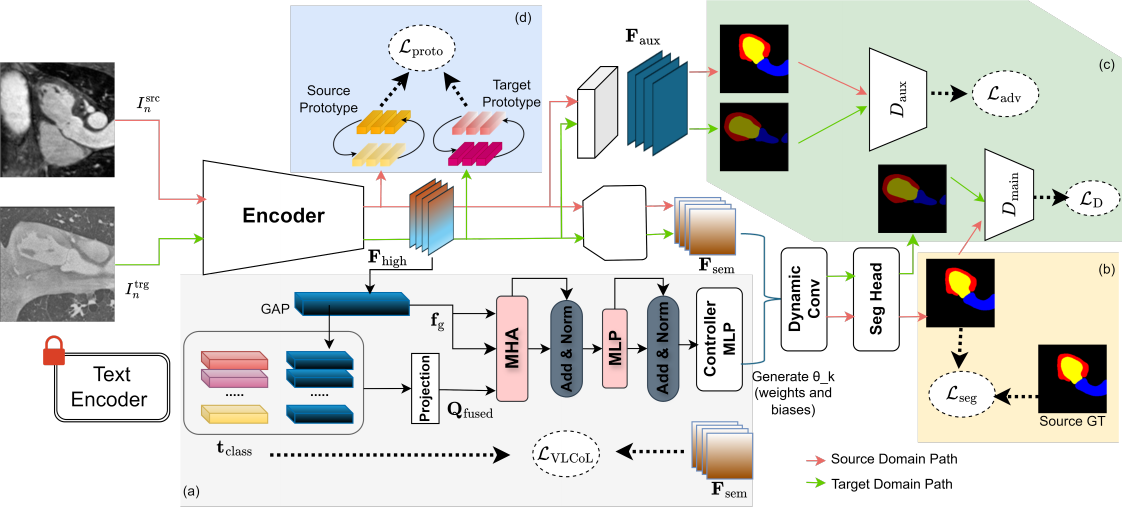}
    \caption{Schematic of the proposed TCSA-UDA framework, comprising: (a) text-driven semantic covariance learning, where modality-aware text embeddings are encoded and used to compute the inter-class text covariance matrix as the alignment target for  $\mathcal{L}_{\text{VLCoL}}$; (b) text-driven supervised learning on the source domain, where visual features are fused with text embeddings via transformer-based fusion and dynamic convolution before being passed to the segmentation head, and $\mathcal{L}_{\text{seg}}$ is computed against source ground truth; (c) adversarial learning using both text-driven and auxiliary predictions, where entropy maps from both domains are fed to discriminators $D_{main}$ and $D_{aux}$, and $\mathcal{L}_{\text{adv}}$ trains the segmentation network to produce target-like entropy distributions; and (d) class-wise cross-domain semantic alignment, where source prototypes (from ground-truth labels) and target prototypes (from pseudo-labels) are maintained with EMA updates and aligned through $\mathcal{L}_{\text{proto}}$. }
    \label{fig:block}
\end{figure*}

\section{Methodology}

\subsection{Problem Definition and Overall Framework}
A key challenge in UDA is the domain shift between the labeled source domain \( I^{\text{src}} \)  and the unlabeled target domain \( I^{\text{trg}} \), which leads to poor generalization of models trained only on \( I^{\text{src}} \). To address this, UDA methods aim to learn domain-invariant representations by jointly leveraging both domains during training. This can be mathematically formulated as \cite{9672690}:
\begin{equation}
\mathcal{F}_{\text{UDA}}: I^{\text{src}} \cup I^{\text{trg}} \rightarrow Y^{\text{src}} \cup Y^{\text{trg}}
\end{equation}
where, \( \mathcal{F}_{\text{UDA}} \) is a mapping function learned using the labeled source domain $\mathcal{D}_S:\{{I^{\text{src}}, Y^{\text{src}}}\}_{i=1}^{N_s}$ and the unlabeled target domain $\mathcal{D}_T: \{I^{\text{trg}}\}_{i=1}^{N_t}$.  Here, each ($i$ th) source and target image $ I_i^\text{src}, I_i^\text{trg} \in \mathbb{R}^{1 \times H \times W}$ and the corresponding source label $Y_i^\text{trg} \in \{1, \dots, C\}^{H \times W}$.  Here, $C$ represents the classes and ${H \times W}$ denotes the size of image. In adversarial learning \cite{tsai2018learning},  a segmentation network \( G \) generates predictions, and a discriminator \( D \) attempts to distinguish whether these predictions originate from the source or target domain.  The network \( G \)  is trained to deceive \( D \) by generating target-like predictions. Although adversarial approaches are effective, conventional UDA frameworks typically assume domain homogeneity \cite{Du_2024_CVPR}. In real-world medical scenarios, datasets often span multiple imaging modalities such as CT, MRI, and ultrasound across both source and target domains. These modalities are often unpaired and heterogeneous,  and can be denoted as \( \{D_{M_1}, D_{M_2}, \dots, D_{M_N}\} \). 

We therefore extend UDA to a cross-modal adaptation setting, aiming to train a single segmentation model capable of generalizing across diverse modalities. To facilitate this, we incorporate text embeddings \( T \in \mathbb{R}^d \), derived from modality-aware class labels using pretrained language models (e.g. CLIP \cite{pmlr-v139-radford21a} or BioBERT \cite{10.1093/bioinformatics/btz682}). These embeddings guide the visual feature learning via cross-attention or projection alignment, providing semantic priors that improve consistency across both domains and modalities. The problem is thus formulated as learning a unified function \( G(I, T) \) that jointly minimizes domain discrepancy and enhances cross-modal semantic alignment without requiring target labels. 

The overall architecture of our proposed approach is illustrated in Fig.  \ref{fig:block} and consists of three key components. First (Fig. \ref{fig:block} (a) and (b)), to address cross-domain medical image segmentation, we propose a modality-aware vision-language fusion framework that integrates visual features with shared semantic class representations derived from modality-aware text prompts through a transformer-based fusion module and dynamic convolution. While modality-specific cues (e.g., CT or MRI) are incorporated at the prompt level to improve vision–language alignment, domain invariance is enforced through downstream semantic alignment, enabling effective generalization across heterogeneous imaging modalities without requiring pixel-level annotations from the target domain. Second (Fig. \ref{fig:block} (c)), we employ two Generative Adversarial Networks (GANs) to perform domain alignment through adversarial learning. The generator produces two outputs: (i) a text-driven segmentation prediction used for supervised learning on the source domain, and (ii) an auxiliary prediction that generates self-information maps for both source and target domains, used to fool domain discriminators (\(D_{\text{main}}\) and \(D_{\text{aux}}\)). This adversarial setup encourages domain-invariant predictions and category-aware adaptation. Third (Fig. \ref{fig:block} (d)), to mitigate residual class-level distribution mismatches that are not fully addressed by entropy-based constraints, we incorporate an adaptive prototype alignment module. This module leverages class-level feature representations and aligns them across domains, facilitating consistent pixel-level feature alignment of corresponding semantic categories between the source and target domains. The detailed implementation of each component is described in the following subsections.

\subsection{Text-Driven Cross Domain Learning}
Text-driven cross domain learning helps the model perform well on different imaging types like CT and MRI. As shown in Fig. \ref{fig:block}, our encoder architecture builds on DeepLabV2 with a ResNet backbone to extract multi-scale features. A parallel text branch provides modality-aware semantic guidance, and a transformer-based fusion module aligns the visual and textual features for better cross-domain understanding. The encoder outputs high-level features $
\mathbf{F}_\text{high} \in \mathbb{R}^{B \times 2048 \times H \times W}$. These features undergo a sequence of convolutions with channel dimensions [1024, 512, 256, 128] to refine visual representations, resulting in semantic feature $
\mathbf{F}_\text{sem} \in \mathbb{R}^{B \times 256 \times H \times W}
$. The classifier uses atrous (dilated) convolutions to capture multi-scale context, produces auxiliary segmentation logits $ \mathbf{F}_\text{aux} \in \mathbb{R}^{B \times C \times H \times W}$, where \( B \) is batch size, \( H \times W \) are the spatial dimensions of the feature maps, and \( C \) is the number of classes.
Semantic text embeddings are generated using modality-aware medical prompts for each organ/class.  Specifically, prompts are constructed using the template  “A \{Dataset\} \{CT/MRI\} imaging of a [CLS]”.  Here, \{CT/MRI\} specifies the imaging modality, either computed tomography or magnetic resonance, \{Dataset\} represents specific medical dataset type such as cardiac or abdominal, and [CLS] is replaced with a class-specific anatomical term (e.g. myocardium of the left ventricle, left atrium blood cavity).  
The prompt template is intentionally designed to incorporate three types of contextual information: the dataset context, the imaging modality, and the anatomical class label. This formulation allows the language encoder to capture both modality-aware contextual information and shared anatomical semantics. While the modality token (e.g., CT or MRI) helps the model adapt to modality-specific visual characteristics, the class token [CLS] serves as a domain-invariant semantic anchor. Consequently, the resulting text embeddings encode both modality context and anatomical semantics, enabling more effective cross-modal feature alignment.
All prompts are encoded using a shared pretrained language model (e.g., CLIP or BioBERT \cite{10.1093/bioinformatics/btz682}), ensuring that semantic relationships between classes are preserved in a common embedding space. For a given input image, the embedding corresponding to its modality,  \( \mathbf{E}_{\text{modality}} \), is selected and projected to a 256-dimensional space to produce $\mathbf{t}_{\text{class}} \in \mathbb{R}^{C \times 256}$, which captures semantic encoding for each class.\par
On the vision side, high-level global features $\mathbf{F}_\text{high}$ are aggregated using global average pooling, followed by group normalization and   a $1\times1$ convolution, to obtain a global visual representation $\mathbf{F}_{\text{GAP}}\in \mathbb{R}^{B \times 256 \times 1 \times 1}$. This feature is flattened to form a global visual feature vector $\mathbf{f}_{\text{g}} \in \mathbb{R}^{B \times 1 \times 256}$. This global feature is repeated $C$ times to match the number of classes, resulting in $\mathbf{f}_{\text{rep}}\in \mathbb{R}^{B \times C \times 256}$. The repeated visual feature is then concatenated with the corresponding $\mathbf{t}_{\text{class}}$, and the concatenated representations are  passed through a linear layer with ReLU to obtain the fused query representation $\mathbf{Q}_{\text{fused}}\in \mathbb{R}^{B \times C \times 256}$. \added{Each of the C rows of $\mathbf{Q}_{\text{fused}}$ represents  a class-specific combination of language semantics and global visual context, enabling selective attention over visual features that is conditioned on both the anatomical class identity and the imaging modality. The fused query is then fed into a Transformer-based  multi-head attention (MHA) module, where the global visual feature $\mathbf{f}_{\text{g}}$ is used as both the key and value. This allows the class-conditioned textual queries to attend to the global visual representation and generate modality-aware visual-language features.}  The multi-layer perceptron (MLP) refines this fused representation, enabling precise modality-aware semantic segmentation.
\begin{equation}
    \mathbf{F}_{\text{fused}} = \text{MLP}(\text{MHA}(\mathbf{Q}_{\text{fused}}, \mathbf{f}_{\text{g}}, \mathbf{f}_{\text{g}})) \in \mathbb{R}^{B \times C \times 256}
\end{equation}
Here, $\mathbf{F}_{\text{fused}}$ encodes class-wise semantic information conditioned on both the imaging modality and the input visual context. To adapt the segmentation features according to the fused visual-language representation, we use a dynamic convolution module. Specifically, the class-wise fused features are averaged along the class dimension and passed to an MLP controller:
\begin{equation}
\small
\theta_k = \text{MLP}(\text{mean}(\mathbf{F}_{\text{fused}}, \text{dim}=1)) \in \mathbb{R}^{B \times (256 \times 128 + 128)}
\end{equation}
These parameters generate dynamic convolution weights $\mathbf{W} \in \mathbb{R}^{B \times 128 \times 256 \times 1 \times 1}$ and biases $\quad \mathbf{b} \in \mathbb{R}^{B \times 128}$ for each input sample. This mechanism allows per-sample convolution filter generation, improving model adaptability to modality-aware features. For each  visual sample $\mathbf{F}_{\text{sem}}$, a customized convolution is applied
\begin{equation}
  \mathbf{f}^{\text{conv}} = \text{Conv2D}(\mathbf{F}_{\text{sem}}, \mathbf{W}, \mathbf{b}) \in \mathbb{R}^{B\times128 \times H \times W}  
\end{equation}
Since the convolution parameters are generated from the Transformer-fused visual-language representation, the segmentation features are adaptively modulated according to the modality-aware anatomical semantic context. Finally, $\mathbf{f}^{\text{conv}}$ is passed to the Seg Head to produce the segmentation mask $P_n^{\text{src}}$. For a source labeled mask $\mathbf{Y}_n^\text{src}$ , the supervised segmentation loss  $\mathcal{L}_{\text{seg}}$ is calculated as
\begin{equation}
\label{seg_loss}
\mathcal{L}_{\text{seg}} = \sum_{n=1}^{N} \left( \mathcal{L}_{\text{CE}}(P_n^{\text{src}}, \mathbf{Y}_n^\text{src}) + \mathcal{L}_{\text{Dice}}(P_n^{\text{src}}, \mathbf{Y}_n^\text{src}) \right)
\end{equation}
where, $P_n^{\text{src}}$ denotes the prediction for the $(n)$-th source image, and $\mathcal{L}_{\text{CE}}$ and $\mathcal{L}_{\text{Dice}}$ denote the cross-entropy and Dice loss functions.  
Importantly, domain invariance is not imposed by identical prompt wording, but is instead enforced through (i) shared class semantics, (ii) a shared text encoder, and (iii) downstream cross-modal alignment losses that map visual features from different domains into a unified semantic space.

\subsection{Vision-Language Covariance Cosine Loss (VLCoL)}

VLCoL enforces that the relationships (covariances) among class-specific pixel features extracted from the segmentation encoder match the relationships among corresponding text embeddings of those classes. The alignment is done by minimizing the cosine distance between the two covariance matrices. This encourages the segmentation encoder’s latent space to reflect the semantic structure embedded in text, improving domain adaptation by leveraging semantic consistency. To extract class-specific pixel features $\mathbf{f}^c$, we use the downsampled ( same as feature shape) ground-truth mask to generate a binary mask $b^c$ for class $c$ and apply spatial averaging:

\begin{equation}
\mathbf{f}^c = \frac{1}{H} \cdot \ \frac{1}W \sum_{i=1}^{H} \sum_{j=1}^{W} b_{i,j}^c \cdot \mathbf{F_\text{sem}}_{i,j}
\end{equation}
where, $b^c \in \{0,1\}$ indicates whether a pixel belongs to class $c$, and $\mathbf{F_\text{sem}}_{i,j}$ is the pixel feature at location $(i,j)$. We then construct covariance matrices for pixel features $\Sigma_p$ and the corresponding semantic text embeddings derived from shared class representations $\Sigma_t$ :

\begin{equation}
\Sigma_p = \left[\mathrm{Cov}(\mathbf{f}^i, \mathbf{f}^j)\right]_{i,j=1}^{C},
\Sigma_t = \left[\mathrm{Cov}(\mathbf{t}^i, \mathbf{t}^j)\right]_{i,j=1}^{C}
\end{equation}
where, $\mathbf{f}^i$ and $\mathbf{t}^i$ denote the feature vectors for class $i$, and $C$ is the total number of classes. Due to typically small batch sizes and the fact that not all classes are present in each mini-batch, we compute $\Sigma_t$ using only the text embeddings corresponding to classes present in the sample. To address high variance in class pixel features, we use a memory bank $\mathbf{f}^c_{\text{current}}$ to store running averages, updated using an exponential decay:

\begin{equation}
\mathbf{f}^c_{\text{new}} = \lambda \cdot \mathbf{f}^c_{\text{current}} + (1 - \lambda) \cdot \mathbf{f}^c
\label{eq:memory_update}
\end{equation}
where $\lambda \in [0,1]$ is the decay coefficient. The updated features $\mathbf{f}^c_{\text{new}}$ are used to compute $\Sigma_p$ during training. The final VLCoL minimizes the cosine distance between the pixel covariance matrix $\Sigma_p$ and the text covariance matrix $\Sigma_t$, encouraging the encoder’s feature space to emulate the semantic structure of the domain-agnostic text space:

\begin{equation}
\mathcal{L}_{\text{VLCoL}} = 1 - \frac{ \langle \Sigma_p, \Sigma_t \rangle }{ \| \Sigma_p \|_F \cdot \| \Sigma_t \|_F }
\end{equation}
where $\langle \cdot, \cdot \rangle$ denotes the Frobenius inner product and $\| \cdot \|_F$ is the Frobenius norm. VLCoL can be integrated into any UDA framework that uses labeled source data and unlabeled target images. Rather than performing source–target distribution matching, VLCoL regularizes the encoder to preserve language‑derived inter‑class structure in its latent space, providing a semantic prior that complements conventional target‑domain alignment losses.

\subsection{Adversarial Learning for Structural Alignment}
Following prior work \cite{vu2019advent}, we enforce alignment of structural entropy distributions by training a discriminator to distinguish between domain-specific entropy maps. Specifically, for a batch of \( N \) labeled source samples \( \{(I_n^{\text{src}}, Y_n^{\text{src}})\}_{n=1}^N \) and \( N \) unlabeled target images \( \{I_n^{\text{trg}}\}_{n=1}^N \), our segmentation network \( G \) generates pixel-wise probability maps \( P_n^{\text{src}} \) and \( P_n^{\text{trg}} \) via the softmax function. We convert these probability maps into entropy maps \( E_n^{\text{src}} \) and \( E_n^{\text{trg}} \) by computing the self-information at each pixel, i.e. the negative product of the probability and its logarithm. These entropy maps are then fed into a discriminator \( D \), which is trained using a binary cross-entropy loss:
\begin{equation}
    \mathcal{L}_{\text{D}} = \sum_{n=1}^N \left( 
\mathcal{L}_\text{BCE}(D(E_n^{\text{src}}), 1) + 
\mathcal{L}_\text{BCE}(D(E_n^{\text{trg}}), 0) \right)
\end{equation}

Meanwhile, the segmentation network (generator) is adversarially trained to fool the discriminator by minimizing:
\begin{equation}
\label{eq_adv}
    \mathcal{L}_{\text{adv}} = \sum_{n=1}^N \mathcal{L}_\text{BCE}(D(E_n^{\text{trg}}), 1)
\end{equation}
This adversarial mechanism encourages the network to produce structurally consistent predictions across domains, effectively reducing entropy discrepancy and improving generalization on the target domain.

\subsection{Prototype Alignment}

From high-level feature $
\mathbf{F}_\text{high} $ extracted from the penultimate layer of the network, we collect pixel-level embeddings \( \mathbf{e}_v^s \) and \( \mathbf{e}_v^t \), where \( v \) denotes a pixel, and the superscripts \( s \) and \( t \) refer to the source and target domains, respectively. For every class \( c \in \{1, 2, \dots, C\} \), we compute class-specific prototypes from both domains. The source domain prototype \( \mathbf{z}_c^s \) is obtained by averaging the feature vectors \( \mathbf{e}_v^s \) over all pixels that are labeled with class \( c \), using the ground-truth label \( y_v^s \). In the target domain, the prototype \( \mathbf{z}_c^t \) is calculated similarly, but using pseudo-labels \( \hat{y}_v^t \), which are derived by taking the class with the maximum predicted softmax probability, i.e. \( \hat{y}_v^t = \arg\max(\mathbf{p}_v^t) \). The set of pixels belonging to class \( c \) in the source and target domains are denoted by \( \Omega_c^s \) and \( \Omega_c^t \), respectively. Following prior prototype-based domain adaptation methods, including Xie et al. \cite{xie2018learning}, we maintain global class prototypes using exponential moving averages to stabilize class-level alignment under mini-batch training. 
The global prototype updates are performed as follows:
\begin{align}
    \mathbf{z}_c^s \leftarrow \beta \mathbf{z}_c^s + (1 - \beta) \cdot \frac{1}{|\Omega_c^s|} \sum_{v \in \Omega_c^s} \delta(y_v^s = c) \cdot \mathbf{e}_v^s\\
\mathbf{z}_c^t \leftarrow \beta \mathbf{z}_c^t + (1 - \beta) \cdot \frac{1}{|\Omega_c^t|} \sum_{v \in \Omega_c^t} \delta(\hat{y}_v^t = c) \cdot \mathbf{e}_v^t
\end{align}
where, \( \beta \in [0, 1] \) is a momentum parameter that controls the influence of the current batch on the running prototype, and \( \delta(\cdot) \) is the indicator function that equals 1 when the condition inside is true, and 0 otherwise. Finally, to enforce the alignment between source and target semantic spaces, we apply a prototype alignment loss. This loss penalizes the distance between corresponding source and target prototypes for each class using the squared \( \ell_2 \)-norm:
\begin{equation}
    \mathcal{L}_{\text{proto}} = \sum_{c=1}^{C} \|\mathbf{z}_c^s - \mathbf{z}_c^t\|^2
\end{equation}
By minimizing \( \mathcal{L}_{\text{proto}} \), the model is encouraged to bring class-level feature representations from both domains closer together, thereby promoting better cross-domain semantic consistency and improving the segmentation performance under domain shift.

In summary, the overall training objective of the proposed segmentation network is defined as:
\begin{equation}
\mathcal{L}_{\text{total}} = \mathcal{L}_{\text{seg}} + \lambda_1 \mathcal{L}_{\text{adv}} + \lambda_2 \mathcal{L}_{\text{VLCoL}} + \lambda_3 \mathcal{L}_{\text{proto}}
\end{equation}
where $\lambda_1$, $\lambda_2$ and $\lambda_3$ are balance coefficients. The segmentation loss $\mathcal{L}_{\text{seg}}$ is computed only on the labeled source images and defined in Eq. \ref{seg_loss}. $\mathcal{L}_{\text{adv}}$ is the adversarial loss for domain alignment, defined in Eq. \ref{eq_adv}.

\section{Experiments and Results}

\subsection{Datasets} 
\subsubsection{Cardiac Substructure Segmentation Dataset} To evaluate the effectiveness of our proposed method, we conduct experiments on the Multi-Modality Whole Heart Segmentation (MMWHS) Challenge 2017 dataset \cite{zhuang2016multi}, a widely used benchmark for cross-modality medical image segmentation. The dataset consists of 20 unpaired cardiac CT volumes and 20 unpaired cardiac MRI volumes, each with ground-truth annotations. These scans are collected from different imaging centers and patient cohorts, introducing significant domain shifts. We focus on segmenting four key cardiac structures: the ascending aorta (AA), left atrium cavity (LAC), left ventricle cavity (LVC), and myocardium of the left ventricle (MYO). Domain adaptation is evaluated in both CT→MRI and MRI→CT directions. 

\subsubsection{Abdominal Organ Segmentation Dataset} We further validate our approach on an abdominal dataset consisting of 30 CT volumes from publicly available source \cite{landman2015miccai} and 20 T2-SPIR MRI volumes from the ISBI 2019 CHAOS Challenge dataset \cite{kavur2021chaos}. Both modalities include pixel-wise annotations for four abdominal organs: Spleen, Right Kidney (RK), Left Kidney (LK), and Liver. We employ the preprocessed dataset provided by SIFA V2 \cite{8988158}.

\subsubsection{Brain Tumor Segmentation Dataset} We also utilized the Multi-Modality Brain Tumor Segmentation Challenge 2018 (BraTS 2018) dataset \cite{menze2014multimodal} for tumor segmentation. This dataset comprises scans from 285 patients, including 210 high-grade glioma (HGG) and 75 low-grade glioma (LGG) cases, with four MRI modalities: T1, T1CE, T2, and FLAIR. For our experiment, we perform bidirectional domain adaptation between the FLAIR and T2 modalities using the 75-subject LGG cohort. To further investigate generalization across tumor grades, we additionally selected a fixed 50\% subset of the HGG cohort, comprising 105 of the 210 available HGG subjects. The HGG subset was evaluated under the same FLAIR→T2 and T2→FLAIR adaptation protocol used for LGG.

\begin{table*}[!ht]
\centering
\caption{Comparison of segmentation performance on the MMWHS dataset for Cardiac MRI $\rightarrow$ CT and CT $\rightarrow$ MRI domain adaptation. }
\setlength{\tabcolsep}{3pt} 
\renewcommand{\arraystretch}{1.2}
\label{tab:combined_mrct_ctmr}
\resizebox{\textwidth}{!}{%
\begin{tabular}{l|ccccc|ccccc||ccccc|ccccc}
\toprule
\multirow{3}{*}{Method} 
& \multicolumn{10}{c||}{\textbf{MRI $\rightarrow$ CT}} 
& \multicolumn{10}{c}{\textbf{CT $\rightarrow$ MRI}} \\
\cmidrule(lr){2-11} \cmidrule(lr){12-21}
& \multicolumn{5}{c|}{Dice ($\uparrow$)} & \multicolumn{5}{c||}{ASD ($\downarrow$)} 
& \multicolumn{5}{c|}{Dice ($\uparrow$)} & \multicolumn{5}{c}{ASD ($\downarrow$)} \\
\cmidrule(lr){2-6} \cmidrule(lr){7-11} \cmidrule(lr){12-16} \cmidrule(lr){17-21}
& AA & LAC & LVC & MYO & Avg & AA & LAC & LVC & MYO & Avg
& AA & LAC & LVC & MYO & Avg & AA & LAC & LVC & MYO & Avg \\
\midrule
Supervised training & 89.3 & 91.4 & 92.8 & 88.0 & 90.4 & 2.3 & 2.9 & 1.5 & 3.2 & 2.5 
& 81.6 & 86.3 & 92.3 & 80.0 & 85.1 & 3.4 & 2.1 & 1.7 & 1.6 & 2.2\\
W/o adaptation & 30.8 & 36.8 & 18.3 & 7.2 & 23.3 & 20.2 & 8.9 & 33.6 & 27.8 & 22.6 
& 18.5 & 7.3 & 53.5 & 2.1 & 20.4 & 7.1 & 25.8 & 8.7 & 29.9 & 17.9\\
\midrule
CycleGAN \cite{8237506} & 73.8 & 75.7 & 52.3 & 28.7 & 57.6 & 11.5 & 13.6 & 9.2 & 8.6 & 10.8 
& 64.3 & 30.7 & 65.0 & 43.0 & 50.7 & 5.8 & 9.8 & 6.0 & 5.0 & 6.6\\
AdaOutput \cite{tsai2018learning} & 73.5 & 80.4 & 76.1 & 48.6 & 69.6 & 15.5 & 5.8 & 5.2 & 6.6 & 8.3 
& 52.3 & 71.1 & 79.5 & 49.2 & 63.2 & 9.0 & 3.5 & 5.1 & 5.4 & 5.8\\
PnP-AdaNet \cite{8764342} & 74.0 & 68.9 & 61.9 & 50.8 & 63.9 & 12.8 & 6.3 & 17.4 & 14.7 & 12.8 
& 43.7 & 47.0 & 77.7 & 48.6 & 54.3 & 11.4 & 14.5 & 4.5 & 5.3 & 8.9\\
AdvEnt \cite{vu2019advent} & 79.5 & 83.0 & 79.5 & 57.7 & 75.0 & 13.9 & 9.3 & 6.9 & 4.5 & 8.7 
& 52.3 & 71.1 & 79.5 & 49.2 & 63.2 & 9.0 & 3.5 & 5.1 & 5.4 & 5.8\\
SIFA V2 \cite{8988158} & 81.3 & 79.5 & 73.8 & 61.6 & 74.1 & 7.9 & 6.2 & 5.5 & 8.5 & 7.0 
& 65.3 & 62.3 & 78.9 & 47.3 & 63.4 & 7.3 & 7.4 & 3.8 & 4.4 & 5.7\\
EBM \cite{9413284} & 78.9 & 80.7 & 75.7 & 60.4 & 74.9 & 8.6 & 6.4 & 4.7 & 8.2 & 7.1 
& 65.9 & 64.2 & 76.9 & 49.1 & 64.0 & 6.9 & 7.5 & 5.6 & 3.8 & 6.0\\
CRST \cite{9010413} & 79.6 & 80.5 & 78.3 & 63.7 & 75.5 & 8.8 & 6.4 & 4.5 & 7.5 & 6.8 
& 65.1 & 66.9 & 77.2 & 50.0 & 64.8 & 6.4 & 6.3 & 5.5 & 4.0 & 5.6\\
DAFormer \cite{9879466} & 85.5 & 88.2 & 74.5 & 60.2 & 77.1 & 12.1 & 9.2 & 7.7 & 4.9 & 8.5 
& \textbf{75.2} & 59.4 & 72.0 & 57.1 & 65.9 & 4.7 & 13.8 & 11.6 & 7.8 & 9.5\\
ADR \cite{sun2022attention} & 87.9 & 86.8 & 82.1 & 64.2 & 80.2 & 5.9 & 4.1 & 4.8 & 5.7 & \textbf{5.1} 
& 66.9 & 69.1 & 81.0 & 48.3 & 66.3 & 6.9 & 5.5 & 3.4 & 4.0 & 4.9\\
SECASA \cite{Feng_Ju_Wang_Song_Zhao_Ge_2023} & 83.8 & 85.2 & 82.9 & 71.7 & 80.9 & 9.6 & 4.2 & 3.9 & 3.9 & 5.4 
& 68.3 & 74.6 & 81.0 & 55.9 & 69.9 & 4.9 & 3.6 & 5.4 & \textbf{3.2} & 4.3\\
MA-UDA \cite{10273225} & \textbf{90.8} & \textbf{88.7} & 77.6 & 67.4 & 81.1 & \textbf{5.7} & \textbf{3.8} & 7.6 & 5.2 & 5.6 
& 71.0 & 67.4 & 77.5 & 59.1 & 68.7 & \textbf{4.4} & 6.9 & 5.6 & 4.2 & 5.3\\
SFDA-CCRC \cite{ma2025source} & 84.1 & 85.6 & 85.2 & 61.8 & 79.2 & 7.2 & 9.5 & \textbf{3.0} & \textbf{3.3} & 5.8 
& 64.3 & \textbf{81.3} & 83.4 & 38.6 & 66.9 & 5.4 & \textbf{3.4} & \textbf{3.0} & 5.1 & \textbf{4.2}\\
IPLC \cite{zhang2024iplc} & 85.6 & 77.2 & 82.4 & 68.4 & 78.4 & 9.6 & 7.5 & 3.9 & 4.3 & 6.3
& 66.3 & 67.9 & 79.9 & 58.2 & 68.1 & 6.5 & 6.6 & 5.2 & 4.5 & 5.7\\
IPLC+ \cite{zhang2025iplc+} & 77.7 & 86.2 & 82.7 & 72.2 & 79.7 & 11.2 & 4.1 & 4.3 & 3.9 & 5.9 
& 64.1 & 71.6 & 80.9 & 59.1 & 68.9 & 4.5 & 6.5 & 3.9 & 3.4 & 4.6\\
\midrule
\textbf{Ours} & 82.5 & 87.1 & \textbf{85.7} & \textbf{74.3} & \textbf{82.4} & 13.2 & 7.2 & 3.5 & 3.6 & 6.9 
& 69.0 & 74.9 & \textbf{83.5} & \textbf{59.2} & \textbf{71.6} & 5.3 & 4.5 & 5.7 & 5.0 & 5.1\\
\bottomrule
\end{tabular}%
}
\end{table*}

\begin{figure*}[ht!]
    \centering
    \includegraphics[width=0.9\linewidth]{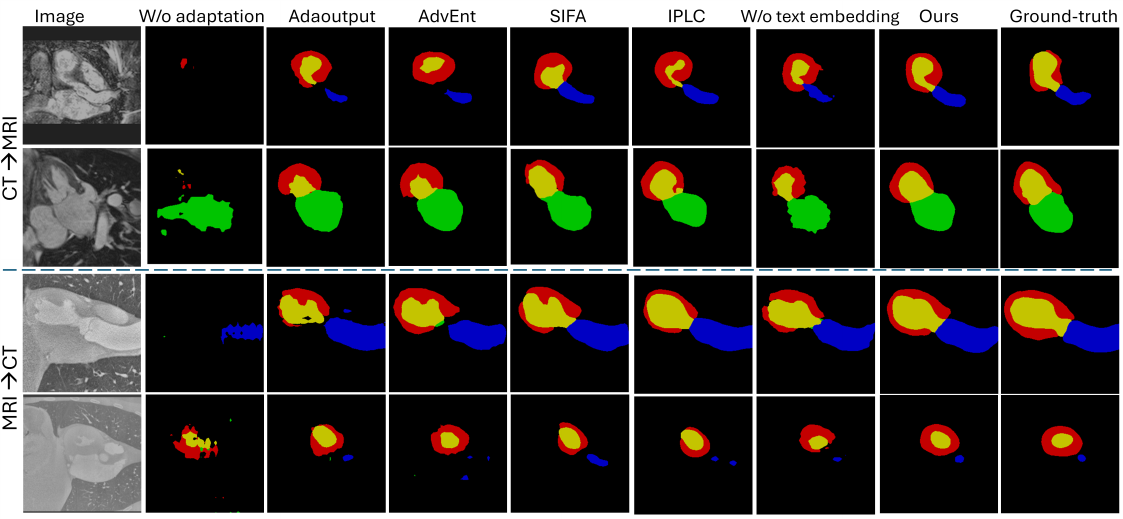}
    \caption{Qualitative segmentation results by different comparison algorithms. Blue, green, yellow, and red represent AA, LAC, LVC, and MYO, respectively.
}
    \label{fig:result}
\end{figure*}

\subsection{Implementation Details} We implement our framework using DeepLabV2~\cite{chen2017deeplab} as the backbone segmentation network $G$, initialized with ImageNet-pretrained weights~\cite{deng2009imagenet}. For adversarial training, we employ a PatchGAN \cite{isola2017image} style discriminator $D$, composed of four convolutional layers followed by a classification layer. All input images are resized to $256 \times 256$ pixels, and standard augmentation techniques—random scaling, rotation, and intensity variation—are applied to improve generalization. The segmentation model is trained using SGD with a learning rate of $2.5 \times 10^{-4}$, momentum $0.9$, and weight decay $5 \times 10^{-4}$, following optimization strategies from~\cite{tsai2018learning, Feng_Ju_Wang_Song_Zhao_Ge_2023}. The discriminator and text fusion module are trained using Adam with a learning rate of $1 \times 10^{-4}$. 
We use a batch size of 4 and set the key hyperparameters as follows: the prototype update coefficient $\beta = 0.01$, and the loss weights $\lambda_1 = 0.003$, $\lambda_2 = 1.0$, and $\lambda_3 = 0.1$ for all translation directions.   All models are implemented in PyTorch and trained on a single NVIDIA A100 GPU with 40 GB memory. For a more robust evaluation, each experiment was conducted with different model initializations, and the average values were used for analysis. 

For the MMWHS cardiac dataset, we adopted the same subject-level train/test partition protocol as used in SIFA V2 \cite{8988158} and subsequent cross-modality UDA studies to ensure direct and fair comparison with previously reported results. For the abdominal and BraTS datasets, we performed repeated evaluations using four different random subject-level 80/20 training/testing splits with a fixed set of random seeds. The final performance was reported as the average across these four splits to reduce the dependence of the results on a single random partition. All data partitions were conducted at the subject level, ensuring that no subject appeared in both the training and testing sets and preventing slice-level data leakage. During training, 2D image slices were randomly selected only from the training subjects and fed into the networks. Segmentation performance is assessed using Dice Similarity Coefficient (Dice) and Average Symmetric Surface Distance (ASD), where higher Dice and lower ASD values indicate better segmentation quality.

\begin{table*}[!p]
\centering
\caption{Comparison of Abdominal Organ Segmentation Performance from Abdominal MRI → CT}
\label{tab:abdominal_mr2ct}
\resizebox{\textwidth}{!}{%
\begin{tabular}{l|ccccc|ccccc}
\toprule
\multirow{2}{*}{Method} & \multicolumn{5}{c|}{Dice (↑)}                                       & \multicolumn{5}{c}{ASD (↓)}                                   \\
\cmidrule(lr){2-6} \cmidrule(lr){7-11}
                        & Spleen     & RK         & LK         & Liver      & Avg        & Spleen    & RK        & LK        & Liver     & Avg       \\
\midrule                        
Supervised              & 84.83 ± 0.60 & 89.95 ± 1.43 & 88.32 ± 2.06 & 91.44 ± 1.30 & 88.64 ± 1.13 & 0.90 ± 0.06 & 0.47 ± 0.06 & 0.59 ± 0.10 & 0.22 ± 0.10 & 0.54 ± 0.04 \\
Adaoutput \cite{tsai2018learning}              & 77.41 ± 0.73 & 77.27 ± 0.54 & 74.30 ± 0.81 & 85.33 ± 0.56 & 78.57 ± 0.54 & 0.94 ± 0.18 & 1.23 ± 0.40 & 1.65 ± 0.13 & 0.86 ± 0.06 & 1.17 ± 0.14 \\
AdvEnt \cite{vu2019advent}                 & 76.82 ± 1.21 & 77.22 ± 0.57 & 74.67 ± 1.00 & 85.25 ± 0.62 & 78.50 ± 0.61 & 1.03 ± 0.17 & 1.11 ± 0.18 & 1.71 ± 0.12 & 0.87 ± 0.05 & 1.18 ± 0.08 \\
SIFA \cite{8988158}                    & 78.75 ± 0.81 & 80.92 ± 0.52 & 78.58 ± 0.58 & 87.84 ± 0.19 & 81.52 ± 0.35 & 1.30 ± 0.07 & 0.74 ± 0.06 & 1.14 ± 0.12 & 0.57 ± 0.06 & 0.94 ± 0.04 \\
SECASA \cite{Feng_Ju_Wang_Song_Zhao_Ge_2023}                 & 75.52 ± 0.35 & 77.07 ± 0.74 & 73.95 ± 0.34 & 85.93 ± 0.77 & 78.62 ± 0.27 & 1.51 ± 0.29 & 1.09 ± 0.14 & 1.69 ± 0.13 & 0.82 ± 0.15 & 1.27 ± 0.06 \\

IPLC \cite{zhang2024iplc}                & 75.35 ± 0.49 & 82.95 ± 0.40 & 78.59 ± 0.38 & 84.27 ± 0.33 & 80.29 ± 0.24 & 1.41 ± 0.12 & 0.73 ± 0.14 & 1.60 ± 0.13 & 0.54 ± 0.07 & 1.07 ± 0.09 \\

IPLC+ \cite{zhang2025iplc+}           & 76.47 ± 0.44 & 83.54 ± 0.33 & 79.30 ± 0.32 & 83.77 ± 0.38 & 80.77 ± 0.17 & 1.33 ± 0.07 & 0.76 ± 0.08 & 1.70 ± 0.13 & 0.55 ± 0.08 & 1.09 ± 0.08  \\

Ours                    & \textbf{81.15 ± 0.49} & \textbf{83.72 ± 0.40} & \textbf{80.35 ± 0.11} & \textbf{88.43 ± 0.39} & \textbf{83.41 ± 0.30} & \textbf{0.88 ± 0.12} & \textbf{0.64 ± 0.06} & \textbf{0.79 ± 0.08} & \textbf{0.52 ± 0.08} & \textbf{0.71 ± 0.07} \\
\bottomrule
\end{tabular}%
}
\end{table*}

\vspace{1em}

\begin{table*}[!p]
\caption{Comparison of Abdominal Organ Segmentation Performance from Abdominal CT→ MRI}
\label{tab:abdominal_ct2mr}
\resizebox{\textwidth}{!}{%
\begin{tabular}{l|ccccc|ccccc}
\toprule
\multirow{2}{*}{Method} & \multicolumn{5}{c|}{Dice (↑)}                                       & \multicolumn{5}{c}{ASD (↓)}                                   \\
\cmidrule(lr){2-6} \cmidrule(lr){7-11} 
                        & Spleen     & RK         & LK         & Liver      & Avg        & Spleen    & RK        & LK        & Liver     & Avg       \\ 
\midrule
Supervised              & 85.60 ± 1.76 & 87.79 ± 0.97 & 87.22 ± 0.70 & 91.06 ± 1.35 & 87.92 ± 0.65 & 1.42 ± 0.13 & 0.25 ± 0.07 & 1.70 ± 0.15 & 0.31 ± 0.02 & 0.92 ± 0.07 \\
AdaOutput \cite{tsai2018learning}              & 58.25 ± 1.06 & 79.42 ± 0.76 & 68.53 ± 0.95 & 89.55 ± 0.52 & 73.94 ± 0.61 & 2.62 ± 0.34 & 0.90 ± 0.13 & 1.77 ± 0.22 & 0.54 ± 0.11 & 1.46 ± 0.14 \\ 
AdvEnt \cite{vu2019advent}                 & 57.62 ± 1.12 & 82.40 ± 0.39 & 68.55 ± 0.42 & 89.57 ± 0.30 & 74.54 ± 0.37 & 2.68 ± 0.18 & 0.65 ± 0.04 & 1.92 ± 0.15 & 0.50 ± 0.08 & 1.44 ± 0.09 \\
SIFA \cite{8988158}                    & 71.67 ± 0.52 & 87.77 ± 0.51 & \textbf{75.57 ± 0.40} & \textbf{90.48 ± 0.26} & 81.37 ± 0.25 & 1.75 ± 0.09 & 0.42 ± 0.05 & \textbf{1.69 ± 0.13} & \textbf{0.39 ± 0.08} & \textbf{1.06 ± 0.07} \\
SECASA \cite{Feng_Ju_Wang_Song_Zhao_Ge_2023}                 & 69.70 ± 1.22 & 82.55 ± 1.32 & 73.17 ± 0.36 & 89.95 ± 0.13 & 78.84 ± 0.64 & 2.43 ± 0.58 & 0.68 ± 0.06 & 1.87 ± 0.22 & 0.53 ± 0.12 & 1.38 ± 0.22 \\
IPLC \cite{zhang2024iplc}         & 73.63 ± 0.44 & 84.92 ± 0.39 & 74.98 ± 0.25 & 89.38 ± 0.33 & 80.73 ± 0.15 & 2.48 ± 0.23 & 0.68 ± 0.11 & 1.74 ± 0.05 & 0.52 ± 0.05 & 1.36 ± 0.08 \\

IPLC+ \cite{zhang2025iplc+}       & 75.10 ± 0.62 & 82.94 ± 0.40 & 75.34 ± 0.34 & 90.20 ± 0.22 & 80.90 ± 0.14 & 2.21 ± 0.06 & 0.61 ± 0.07 & \textbf{1.69 ± 0.03} & 0.42 ± 0.05 & 1.23 ± 0.04 \\

Ours                    & \textbf{77.30 ± 0.49} & \textbf{89.32 ± 0.55} & 75.52 ± 0.44 & 90.31 ± 0.42 & \textbf{83.11 ± 0.29} & \textbf{1.72 ± 0.12} & \textbf{0.40 ± 0.08} & 1.71 ± 0.08 & 0.41 ± 0.07 & \textbf{1.06 ± 0.06} \\
\bottomrule
\end{tabular}%
}
\end{table*}

\begin{figure*}[!p]
    \centering
    \includegraphics[width=0.9\linewidth]{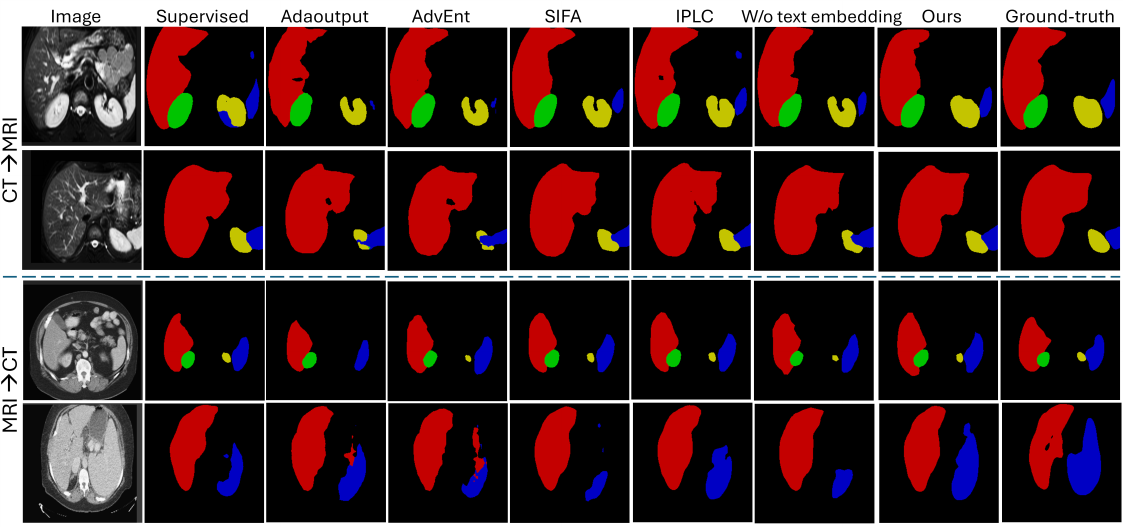}
    \caption{Qualitative results of Abdominal organ segmentation by different comparison algorithms. The blue, green, yellow and red represent the Spleen, RK, LK and Liver,
respectively}
    \label{fig:abdominal_result}
\end{figure*}

\subsection{Comparison with Existing UDAs}
We evaluated the performance of our TCSA-UDA against existing domain adaptation methods, including image-level, feature-level, output-level, transformer-based, semantic-aware and source-free approaches, as detailed in  Table \ref{tab:combined_mrct_ctmr}. For fairness and consistency, we adopted the reported experimental settings from the original works wherever available. Methods lacking public implementations were either re-implemented using the same configuration as SIFA \cite{8988158} or their published results were cited directly for comparison. 
\begin{figure*}[!p]
    \centering
    \includegraphics[width=0.9\linewidth]{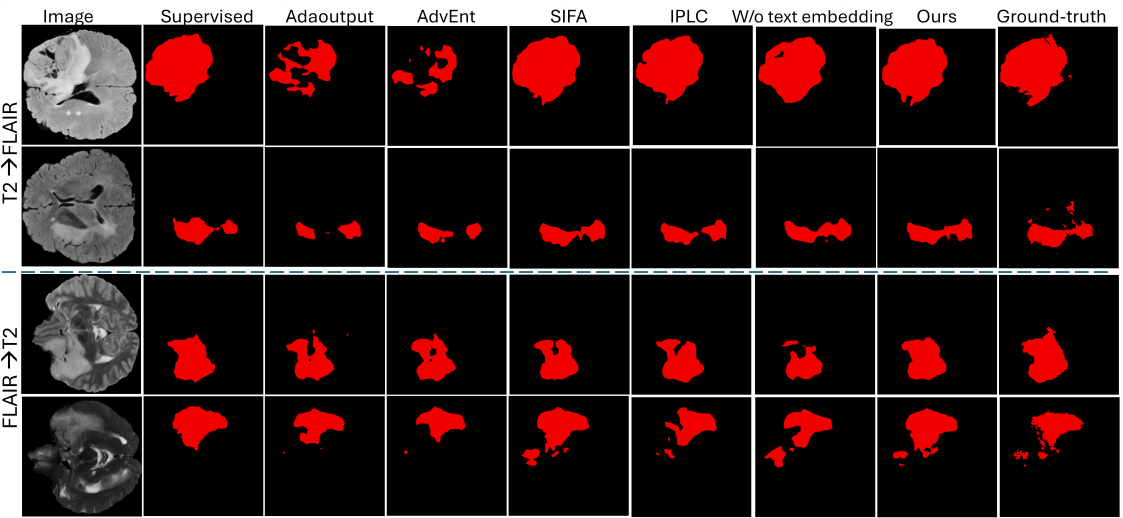}
    \caption{Qualitative results of brain tumor segmentation by different comparison algorithms}
    \label{fig:brats_result}
\end{figure*}

\subsubsection{Results from Cardiac Substructure Segmentation}
For the MRI→CT transfer task, Table \ref{tab:combined_mrct_ctmr} shows that the baseline model without adaptation performs poorly, achieving an average Dice score of only 23.3, while the upper-bound supervised training yields 90.4. Image-level methods like CycleGAN and SIFA V2 focus on style translation but suffer from semantic inconsistency. Our method notably improves CT→MRI Dice (50.7 to 71.6) and MRI→CT Dice (57.6 to 82.4), with better ASD values, demonstrating stronger anatomical fidelity. Feature-level methods such as CRST, EBM and ADR align latent spaces, yet fall short in structure precision. Our Dice scores exceed CRST and EBM in both directions for example, MRI→CT Dice improves from 74.9 (EBM) to 82.4, while maintaining lower ASD (6.9 vs. 7.1). Output-level methods, including AdvEnt and AdaOutput, act on prediction maps but lack structural constraints. Our Dice (CT→MRI: 71.6 vs. AdaOutput’s 63.2) and ASD (5.1 vs. 5.8) reflect stronger regularization via combined semantic and feature guidance. Transformer-based models, such as DAFormer and MA-UDA, leverage global semantics. MA-UDA adds meta-adaptation to dynamically tune representations for each domain. Our method achieves higher CT→MRI Dice (71.6 vs. MA-UDA’s 69.4) and MRI→CT Dice (82.4 vs. DAFormer’s 77.1), highlighting superior cross-domain generalization. Lastly, semantic-aware methods integrate class-level cues; SECASA is competitive with 80.9 Dice (MRI→CT), but our model surpasses this (82.4) while holding steady on ASD. CT→MRI Dice (71.6 vs. 69.9 in SECASA) also favors our method, showing that semantic representation is enhanced without sacrificing boundary precision. Recent source-free and foundation-model-assisted adaptation methods, including SFDA-CCRC , IPLC , and IPLC+ , also improve cross-modality adaptation; however, TCSA-UDA achieves higher average Dice in both MRI→CT (82.4 vs. 79.2, 78.4, and 79.7) and CT→MRI (71.6 vs. 66.9, 68.1, and 68.9), indicating stronger semantic transfer. Our method significantly reduces domain shift compared to the non-adapted baseline, closing 88.1\% of the Dice gap for MRI→CT adaptation and 79.1\% for CT→MRI. Statistical significance testing (p $<$ 0.01) was performed between our method and implemented state-of-the-art approaches, confirming the reliability of its improvements across both MRI→CT and CT→MRI tasks. Figure \ref{fig:result} showcases slice images to illustrate the effectiveness of our method in cardiac structure segmentation across CT and MRI domains. Our approach delivers more accurate and consistent results compared to baseline and UDA methods. By effectively addressing domain shift and preserving both edge sharpness and foreground integrity, the segmented structures closely match the ground truth in both MRI→CT and CT→MRI directions. These consistent improvements across both transfer directions validate the robustness of our approach in handling domain shifts, delivering enhanced segmentation accuracy and spatial coherence for cross-modality cardiac image analysis.

\subsubsection{Results from Abdominal Organ Segmentation} Table \ref{tab:abdominal_mr2ct} and Table \ref{tab:abdominal_ct2mr} show the comparative performance of abdominal organ segmentation, where the supervised model serves as the upper bound and our proposed TCSA-UDA achieves significant improvements over existing unsupervised domain adaptation methods. For MRI → CT, TCSA-UDA attains an average Dice of 83.41\% and ASD of 0.71 mm, outperforming SIFA and SECASA. For CT → MRI, it achieves 83.11\% Dice and 1.06 mm ASD, again surpassing competing approaches. These consistent gains highlight TCSA-UDA’s effectiveness in bridging domain gaps and enhancing cross-modality segmentation accuracy across different anatomical structures and imaging modalities.

Additionally, our experiments show that segmentation methods perform significantly better on abdominal organs than cardiac structures due to anatomical and imaging differences. Abdominal organs are spatially distinct, easing boundary detection, while cardiac structures are closely intertwined, complicating segmentation. Moreover, by leveraging semantic information from textual descriptions of organ classes, the model gains a richer understanding of each structure’s contextual and anatomical characteristics. Fig. \ref{fig:abdominal_result} highlights our method’s effectiveness in abdominal segmentation under bidirectional domain adaptation, producing anatomically plausible and semantically consistent results. Notably, it achieves superior boundary delineation, especially for the liver and spleen, demonstrating robustness in capturing structural and contextual cues where organ separation is clear and modality differences are moderate.

\begin{table}[!p]
\caption{Comparative results of bidirectional adaptation on the BraTS 2018 LGG cohort}
\label{tab_brats_result}
\resizebox{\linewidth}{!}{%
\begin{tabular}{l|cc||cc}
\toprule
\multirow{2}{*}{Method} 
& \multicolumn{2}{c||}{\textbf{FLAIR $\rightarrow$ T2}} 
& \multicolumn{2}{c}{\textbf{T2 $\rightarrow$ FLAIR}}\\
\cmidrule(lr){2-3} \cmidrule(lr){4-5}     
           & Dice (↑)    & ASD (↓)    & Dice (↑)     & ASD (↓)   \\
\midrule
Supervised & 80.72 ± 1.42  & 5.10 ± 0.71& 75.16 ± 1.56  & 5.06 ± 1.13 \\
AdaOutput \cite{tsai2018learning} & 67.56 ± 1.21 & 6.63 ± 1.05  & 67.73 ± 1.71  & \textbf{5.36 ± 1.71} \\
AdvEnt \cite{vu2019advent}    & 67.32 ± 1.32 & 6.74 ± 1.16  & 65.84 ± 2.56  & 5.83 ± 1.94 \\
SIFA \cite{8988158}       & 71.50 ± 0.95 & \textbf{6.62 ± 0.42}  & 72.71 ± 1.46  & 5.67 ± 1.23 \\ 
SECASA \cite{Feng_Ju_Wang_Song_Zhao_Ge_2023}    & 68.27 ± 1.08 & 6.92 ± 1.20  & 70.97 ± 1.51  & 5.64 ± 1.83 \\

IPLC \cite{zhang2024iplc}       & 68.85 ± 1.24 & 7.01 ± 1.34 & 71.03 ± 1.32 & 5.73 ± 1.48 \\

IPLC+ \cite{zhang2025iplc+}      & 69.24 ± 1.04 & 6.87 ± 1.46 & 71.65 ± 1.24 & 5.52 ± 1.27\\

Ours       & \textbf{73.08 ± 0.92} & 7.20 ± 0.26  & \textbf{73.58 ± 0.61}  & 5.90 ± 1.07  \\
\bottomrule
\end{tabular}%
}
\end{table}

\subsubsection{Results from Brain Tumor Segmentation} Table \ref{tab_brats_result} presents quantitative results for bidirectional brain tumor segmentation across FLAIR → T2 and T2 → FLAIR adaptations. For FLAIR → T2, our method achieves a Dice score of 73.08±0.92, outperforming unsupervised baselines such as AdaOutput (67.56 ± 1.21), AdvEnt (67.32 ± 1.32), SECASA (68.27±1.08), and SIFA (71.50 ± 0.95), with a competitive ASD of 7.20 ± 0.26. For T2 → FLAIR, we achieve 73.58±0.61 Dice and 5.90 ± 1.07 ASD, surpassing other unsupervised methods.  Fig. \ref{fig:brats_result} shows the qualitative segmentation results. By enforcing global semantic class consistency through VLCoL and prototype alignment, our method segments a larger proportion of the tumor volume, leading to improved Dice scores. However, minor surface irregularities are observed along tumor boundaries, particularly in the FLAIR→T2 setting, where tumor boundaries are highly heterogeneous and irregular. This explains why the method achieves higher volumetric overlap while showing a slightly higher ASD: it captures more of the tumor mass overall, but occasional boundary roughness increases the average surface distance. Therefore, Dice and ASD show complementary behavior: the model better captures tumor volume, while boundary smoothness remains an area for improvement. 

\begin{figure}
    \centering
    \vspace{1em}
    \includegraphics[width=0.9\columnwidth]{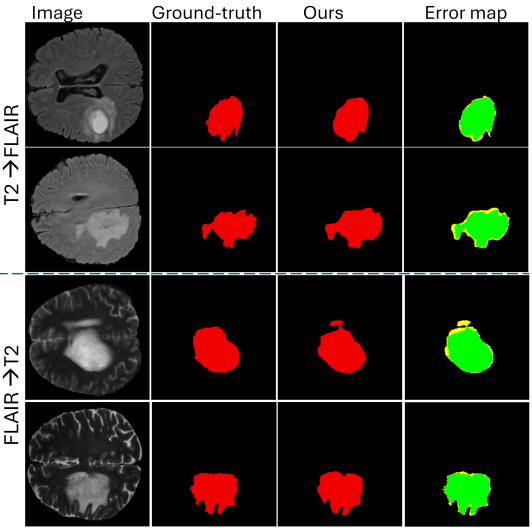}
    \caption{Qualitative evaluation of HGG cases on BraTS 2018. The segmentation error maps show correctly predicted foreground pixels in green and incorrectly predicted pixels, including false positives and false negatives, in yellow. }
    \label{fig:brats_hgg}
    \vspace{1em}
\end{figure}

To further evaluate generalization to more complex tumor morphology, we additionally validated TCSA-UDA on a 50\% subset of the HGG cases from BraTS 2018. As shown in Table \ref{tab:hgg}, TCSA-UDA achieved 74.63 ± 0.89 Dice and 5.58 ± 1.12 ASD for HGG under FLAIR→T2 adaptation. Relative to LGG, this represents a Dice improvement of 1.55 percentage points and an ASD reduction of 1.62. For T2→FLAIR adaptation, HGG achieved 81.54 ± 0.76 Dice and 3.72 ± 0.53 ASD, improving Dice by 7.96 percentage points and reducing ASD by 2.18 compared with LGG. These results indicate that TCSA-UDA maintains comparable volumetric overlap on HGG relative to LGG, while boundary accuracy (ASD) is moderately reduced, consistent with the greater heterogeneity of HGG tumor margins as shown in Fig. \ref{fig:brats_hgg}. The stronger performance on the selected HGG cohort may be attributed to the generally larger and more visually conspicuous tumor regions, which are easier to localize consistently across MRI modalities. Smaller, poorly defined LGG lesions make overlap and boundary metrics more sensitive to minor errors. However, HGG cases still show errors along heterogeneous and irregular margins. Overall, the method transfers effectively across tumor grades without grade-specific tuning by maintaining semantic consistency through VLCoL and prototype alignment. Notably, the segmentation maps produced by our method demonstrate improved preservation of the tumor shape and boundary structure compared with the baseline methods. 
\begin{table}[h]
\caption{Grade-specific evaluation of TCSA-UDA on the BraTS 2018 LGG subset and the fixed 50\% HGG subset.}
\label{tab:hgg}
\resizebox{\columnwidth}{!}{%
\begin{tabular}{@{}lllll@{}}
\toprule
Tumor grade & Cohort size & Adaptation   direction & Avg Dice ↑  & Avg ASD ↓  \\ \midrule
LGG         & 75          & FLAIR→T2               & 73.08 ± 0.92 & 7.20 ± 0.26 \\
HGG         & 105         & FLAIR→T2               & 74.63 ± 0.89 & 5.58 ± 1.12 \\
LGG         & 75          & T2→FLAIR               & 73.58 ± 0.61 & 5.90 ± 1.07 \\
HGG         & 105         & T2→FLAIR               & 81.54 ± 0.76 & 3.72 ± 0.53 \\ \bottomrule
\end{tabular}%
}
\end{table}


\subsection{Ablation Study} 
The ablation analysis is divided into complementary groups to avoid conflating loss-level, representation-level, and architecture-level effects. Table \ref{tab:ablation1} shows the progressive contribution of the principal adaptation objectives, beginning with the visual adversarial baseline and adding prototype alignment and VLCoL on the MRI $\rightarrow$ CT domain adaptation task. The baseline setup includes the segmentation loss ($L_{seg}$) and adversarial loss ($L_{adv}$), achieving an average Dice score of 74.7\%. Incorporating the prototype alignment loss ($L_{proto}$) improves performance to 77.1\%, indicating that semantic alignment between domains enhances segmentation. Including baseline  with the proposed visual-language covariance cosine loss ($L_{VLCoL}$) yields a significant boost to 81.2\%, demonstrating its stronger capacity to guide domain adaptation by leveraging visual-language consistency. When both $L_{proto}$ and $L_{VLCoL}$ are jointly applied alongside $L_{seg}$ and $L_{adv}$, the model achieves the best performance with an average Dice score of 82.4\%. These results highlight the complementary nature of prototype-based alignment and visual-language covariance learning, and confirm that their integration leads to more robust cross-domain feature learning and improved segmentation accuracy. 

\begin{figure}
    \centering
    \vspace{1em}
    \includegraphics[width=1\linewidth]{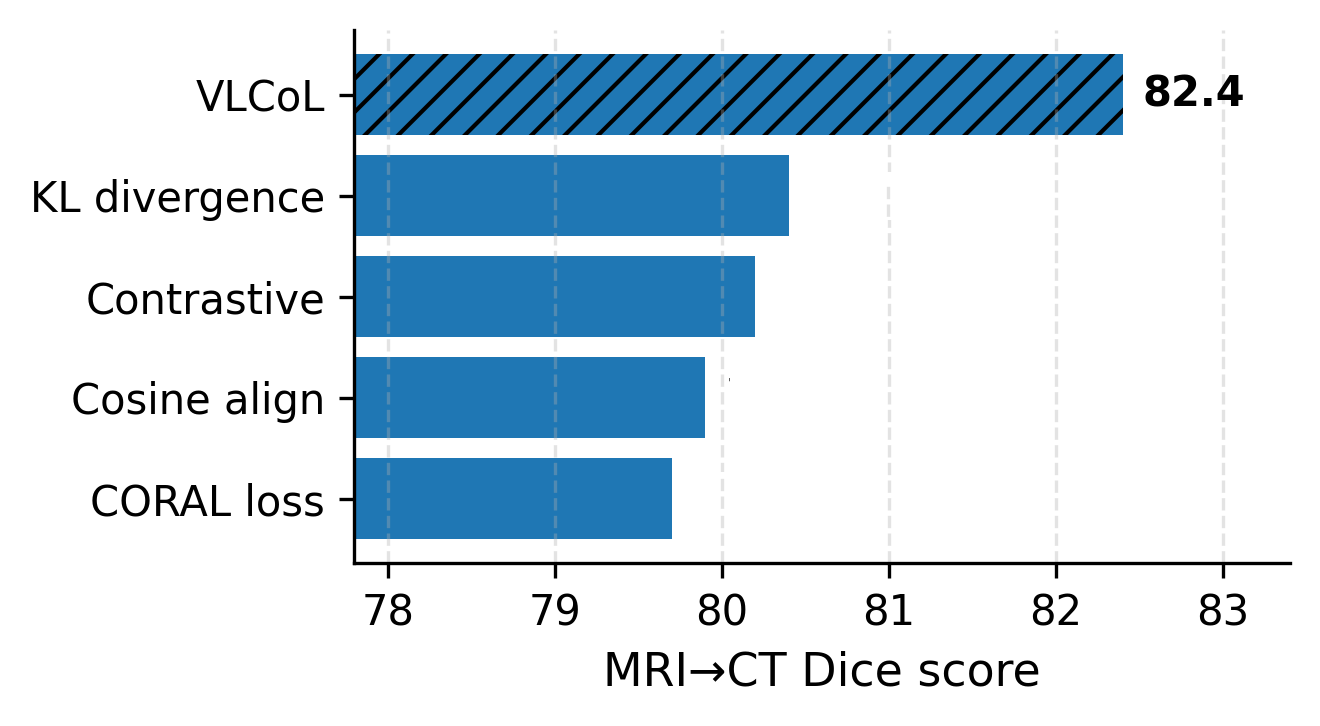}
    \caption{Comparison of alignment objectives for MRI→CT adaptation, including direct cosine prototype-text alignment, contrastive text-image pairing, KL-based relation distribution matching, visual-only CORAL covariance alignment, and the proposed VLCoL alignment. }
    \label{fig:alignment}
    \vspace{1em}
\end{figure}

We further compare different alignment losses for MRI → CT adaptation while keeping the remaining framework unchanged as shown in Fig. \ref{fig:alignment}. The evaluated losses include cosine alignment, which directly matches class visual prototypes with text vectors; contrastive loss, which aligns positive class text-image pairs while treating other classes as negatives; KL divergence, which matches visual and textual relation distributions;  and CORAL loss, which aligns source-target visual covariance statistics without text guidance.  VLCoL achieves the highest MRI→CT Dice score by aligning visual and textual class-relation matrices, suggesting that relation-level vision-language consistency provides a more discriminative and semantically meaningful adaptation signal. 
\begin{table}[!p]

\caption{Ablation study of key components and their loss optimization on the adaptation task for MRI $\rightarrow$ CT. p-values are computed relative to the baseline.}
\renewcommand{\arraystretch}{1.2}
\label{tab:ablation1}
    \centering 
    \resizebox{1\linewidth}{!}{%
    \begin{tabular}{l|p{4mm}p{4mm}p{6mm}p{11mm}|c|c}
        \toprule
        Method & $\mathcal{L}_{seg}$ & $\mathcal{L}_{adv}$ & $\mathcal{L}_{proto}$ & $\mathcal{L}_{VLCoL}$ & Avg Dice & p-value \\
        \midrule
        \midrule
        Baseline &  ~~\checkmark  & ~~\checkmark &            &           & 74.7 & - \\
        + PA only& ~~\checkmark & ~~\checkmark & ~~\checkmark &           & 77.1  & p$<$0.0001\\
        + Text-driven only &  ~~\checkmark & ~~\checkmark &            & ~~~\checkmark & 81.2 & p$<$0.0001\\
  
        \textbf{Ours} & ~~\checkmark & ~~\checkmark & ~~\checkmark & ~~~\checkmark & \textbf{82.4} & p$<$0.0001\\
        \bottomrule
    \end{tabular}
    }
\end{table}

To evaluate the effectiveness of our proposed modality-aware vision-language fusion approach, we conducted a comprehensive ablation study focusing on two key factors: (1) the impact of different text embedding methods and prompt templates along with modality and dataset choice combination, and (2) the role of transformer-based fusion in aligning vision and language representations. As shown in Table~\ref{tab:ablation3}, models utilizing textual embeddings consistently outperformed the vision-only baseline, highlighting the benefit of incorporating semantic priors from language. Among various embeddings, our CLIP-based modality-aware prompt, "A cardiac {CT/MRI} imaging of a [CLS]", achieved the best performance, with 82.4\% and 71.6\% average Dice scores for MRI→CT and CT→MRI adaptation, respectively. These results demonstrate that carefully crafted, modality-aware prompts provide more contextually relevant semantic guidance, leading to better domain adaptation.

Further analysis in Table~\ref{tab:ablation4} emphasizes the complementary role of transformer-based fusion. While shared prompts with fusion improved performance over the non-fused version, combining modality-aware text prompts with transformer fusion and dynamic convolution  yielded the highest gains, outperforming all other configurations. Notably, 'w/o transformer fusion' refers to configurations that exclude MHA and MLP layers from the architecture. The shared text prompt used for all shared configurations is 'A photo of [CLS].' 
\begin{table}[h]

\caption{Ablation study of different embedding methods with their prompt templates for cardiac MRI $\rightarrow$ CT and CT $\rightarrow$ MRI adaptation task}
\label{tab:ablation3}
\resizebox{1\linewidth}{!}{%
\begin{tabular}{@{}p{15mm}|l|p{9mm}p{9mm}|p{10mm}|p{10mm}@{}}
\toprule
Embedding & Prompt   Template & Modality   Token & Dataset   Token & Avg. Dice↑   (MRI $\rightarrow$ CT) & Avg. Dice↑   (CT $\rightarrow$ MRI) \\ \midrule
Vision-Only & —                                      & \xmark & \xmark & 74.7 & 64.5 \\
CLIP        & A photo of   [CLS].                    & \xmark & \xmark & 78.5 & 68.3 \\
CLIP        & A [CLS] in   medical imaging.          & \xmark & \xmark & 78.8 & 68.6 \\
CLIP        & A CT/MRI   imaging of a [CLS].         & \cmark & \xmark & 81.4 & 70.5 \\
CLIP        & A cardiac   imaging of a [CLS].        & \xmark & \cmark & 78.9 & 69.7 \\
\textbf{CLIP (Ours)} & \textbf{A cardiac   CT/MRI imaging of a [CLS]}. & \cmark & \cmark & \textbf{82.4} & \textbf{71.6} \\
BioBERT     & A CT/MRI   imaging of a [CLS].         & \cmark & \xmark & 81.3 & 69.9 \\
BioBERT     & A cardiac   CT/MRI imaging of a [CLS]. & \cmark & \cmark & 81.8 & 70.3 \\ \bottomrule
\end{tabular}%
}
\vspace{1em}

\caption{Ablation experiment results for the combination of modality-aware prompt and transformer fusion.}
\label{tab:ablation4}
\setlength{\tabcolsep}{6pt}
\renewcommand{\arraystretch}{1.15}
\resizebox{\linewidth}{!}{%
\begin{tabular}{l|p{10mm}|p{10mm}|p{10mm}}
\toprule
Methods
& Dynamic conv. 
& Avg. Dice $\uparrow$ MRI$\rightarrow$CT 
& Avg. Dice $\uparrow$ CT$\rightarrow$MRI \\
\midrule
Vision-only  & -- 
& 74.7 
& 64.5 \\

\midrule
Shared prompt + w/o transformer fusion & \xmark 
& 77.6 
& 65.6  \\

Shared prompt + with transformer fusion & \xmark 
& 78.1 
& 67.7  \\

Shared prompt + with transformer fusion & \cmark 
& 78.5 
& 68.3  \\

\midrule
modality-aware prompt + w/o transformer fusion & \xmark 
& 77.6 
& 65.7  \\

modality-aware prompt + with transformer fusion & \xmark 
& 79.8 
& 69.2 \\

\rowcolor{gray!12}
\textbf{modality-aware prompt + with transformer fusion} & \cmark
& \textbf{82.4}
& \textbf{71.6} \\
\bottomrule
\end{tabular}
}
\end{table}

\subsection{Sensitivity of Loss Coefficients.}
\begin{figure}
    \centering
    \vspace{1em}
    \includegraphics[width=1\linewidth]{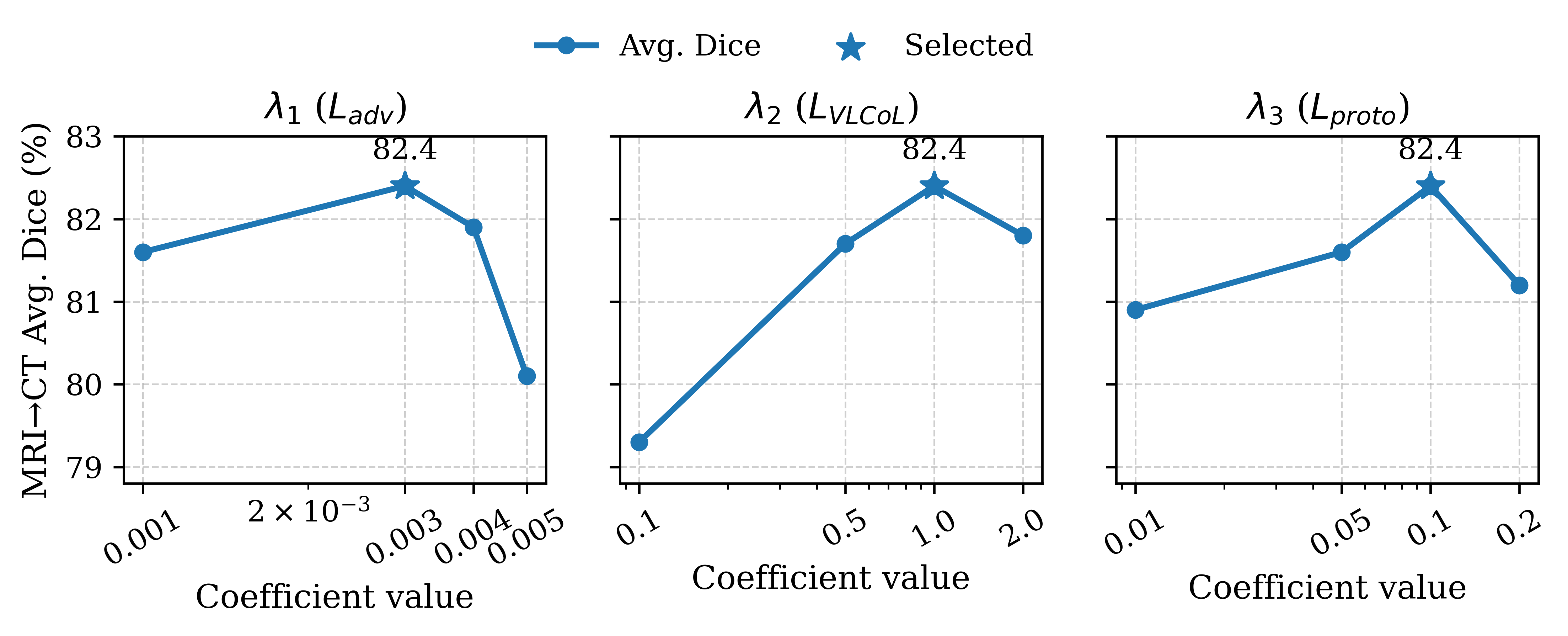}
    \caption{Sensitivity analysis of the loss balancing coefficients on the MRI$\rightarrow$CT task. The selected values \(\lambda_1=0.003\), \(\lambda_2=1.0\), and \(\lambda_3=0.1\) achieve the best average Dice score of \(82.4\%\).}
    \label{fig:hyper}
    \vspace{1em}
\end{figure}

The adversarial loss weight ($\lambda_1$) was selected to be small, following common practice in adversarial UDA segmentation, where the adaptation objective is used as a regularizer rather than the dominant training signal. A large adversarial weight may introduce unstable gradients and force the model to over-align source and target outputs at the expense of source segmentation accuracy. The weights of VLCoL and prototype alignment, ($\lambda_2$) and ($\lambda_3$), are specific to the proposed framework and were selected empirically through sensitivity analysis. Fig.~\ref{fig:hyper} presents the sensitivity analysis of the balancing coefficients \(\lambda_1\), \(\lambda_2\), and \(\lambda_3\) on the MRI\(\rightarrow\)CT adaptation task. For each experiment, one coefficient is varied while the remaining coefficients are fixed at their selected values. The best performance is achieved with \(\lambda_1=0.003\), \(\lambda_2=1.0\), and \(\lambda_3=0.1\), yielding an average Dice score of \(82.4\%\). The results show that the proposed framework remains relatively stable around the selected values, while overly large or small coefficients reduce performance. In particular, a large adversarial weight can overemphasize domain alignment and weaken segmentation discrimination, whereas insufficient or excessive VLCoL/prototype weights may underuse semantic regularization or introduce noisy alignment. These results demonstrate that the selected coefficients provide a balanced contribution from adversarial learning, text-guided covariance alignment, and prototype alignment. The selected coefficients were fixed across all datasets and adaptation directions unless otherwise stated. 

\begin{figure}
    \centering
    \vspace{1em}
    \vspace{1em}
    \includegraphics[width=1\linewidth]{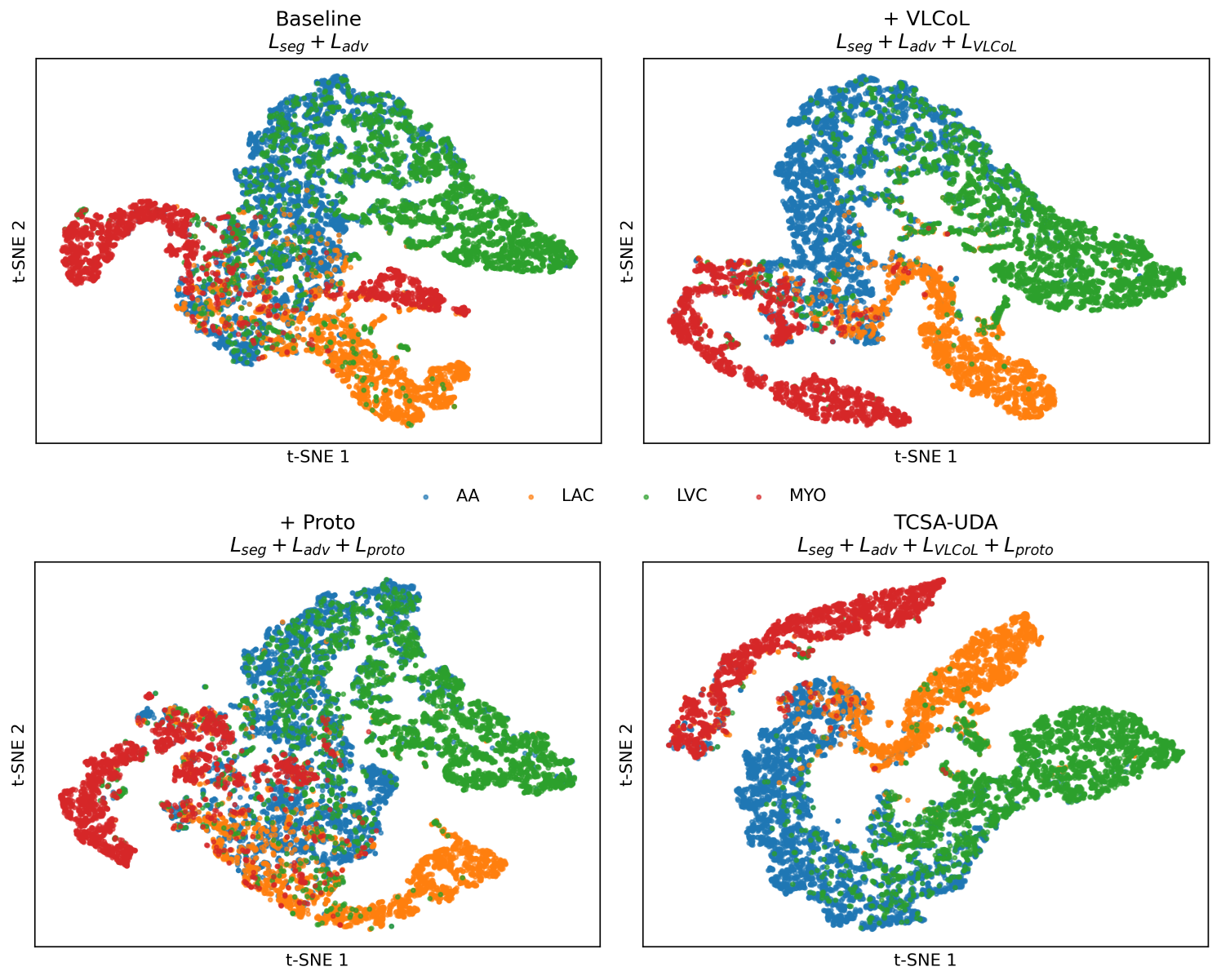}
    \vspace{1em}
    \caption{t-SNE visualization of class-wise feature distributions for the MRI$\rightarrow$CT adaptation task under different training objectives. The baseline model trained with $\mathcal{L}_{seg}+\mathcal{L}_{adv}$ shows considerable inter-class overlap. Adding VLCoL improves semantic separation by aligning visual feature relationships with text-derived class relations, while prototype alignment encourages class-wise consistency. The full TCSA-UDA model, trained with $\mathcal{L}_{seg}+\mathcal{L}_{adv}+\mathcal{L}_{VLCoL}+\mathcal{L}_{proto}$, produces the most compact and well-separated feature clusters, indicating improved intra-class compactness and inter-class discriminability. Different colors denote anatomical classes: AA, LAC, LVC, and MYO.}
    \label{fig:tsne}
    
    \vspace{1em}
\end{figure}

\subsection{T-SNE Visualization}

Fig.~\ref{fig:tsne} visualizes the t-SNE feature distributions for the MRI$\rightarrow$CT adaptation task. The baseline model trained with $\mathcal{L}_{seg}+\mathcal{L}_{adv}$ shows substantial overlap among different anatomical classes, indicating limited class discriminability under cross-modality domain shift. This is also reflected by its low Silhouette score of $0.1605$ and relatively high Davies-Bouldin index of $1.6124$. After introducing VLCoL, the class-wise feature clusters become more compact and better separated, with the Silhouette score increasing to $0.3225$ and the Davies--Bouldin index decreasing to $1.1387$. This indicates that text-derived semantic relation alignment helps organize the visual feature space more discriminatively.

When using prototype alignment alone, the Silhouette score is $0.1034$ and the Davies-Bouldin index is $1.8197$, suggesting that prototype alignment by itself is less effective in producing well-separated feature clusters, likely due to the influence of noisy target pseudo-labels. In contrast, the full TCSA-UDA model, trained with $\mathcal{L}_{seg}+\mathcal{L}_{adv}+\mathcal{L}_{VLCoL}+\mathcal{L}_{proto}$, achieves the best feature organization, with the highest Silhouette score of $0.3416$ and the lowest Davies-Bouldin index of $1.0826$. These results demonstrate that VLCoL and prototype alignment are complementary: VLCoL provides text-guided semantic structure, while prototype alignment further refines class-wise source-target consistency, leading to more compact and discriminative feature representations.


\subsection{GradCAM Visualization Analysis}
\begin{figure*}[!t]
    \centering
    \includegraphics[width=0.8\linewidth]{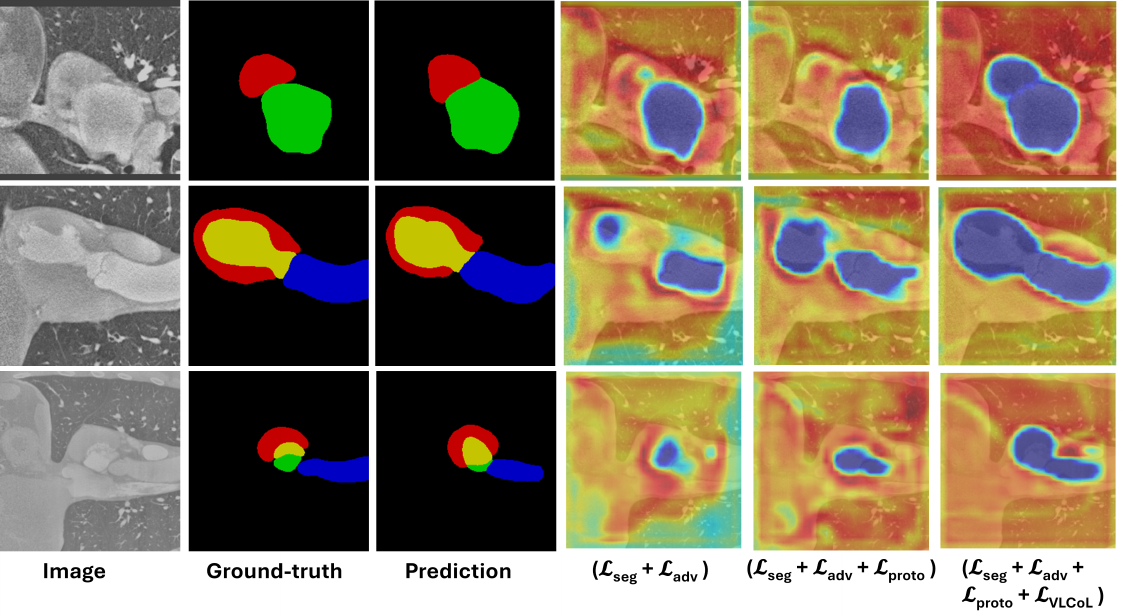}
    \caption{GradCAM visualization results comparing attention regions across different loss configurations in the TCSA-UDA framework. From left to right: input image, ground-truth segmentation, predicted segmentation, and GradCAM maps for models trained with progressively enhanced loss functions—$\mathcal{L}_{\text{seg}} + \mathcal{L}_{\text{adv}}$, $\mathcal{L}_{\text{seg}} + \mathcal{L}_{\text{adv}} + \mathcal{L}_{\text{proto}}$, and $\mathcal{L}_{\text{seg}} + \mathcal{L}_{\text{adv}} + \mathcal{L}_{\text{proto}} + \mathcal{L}_{\text{VLCoL}}$.}
    \label{fig:grad_cam}
\end{figure*}

The GradCAM visualizations in Figure~\ref{fig:grad_cam} provide a detailed view of how attention regions evolve with progressive loss optimization in the TCSA-UDA (Text-Driven Cross-Semantic Alignment for Unsupervised Domain Adaptation) framework. Initially, with the basic combination of segmentation and adversarial losses ($\mathcal{L}_{\text{seg}} + \mathcal{L}_{\text{adv}}$), the attention maps are broad and loosely aligned with semantic regions, indicating limited semantic precision despite domain alignment. The introduction of the prototype loss ($\mathcal{L}_{\text{proto}}$) enhances class-wise feature consistency, resulting in more focused and semantically meaningful attention. The most significant refinement is observed with the inclusion of the vision-language covariance cosine loss ($\mathcal{L}_{\text{VLCoL}}$), which leverages cross-modal covariance to guide the model’s attention toward contextually relevant and semantically rich regions. This loss encourages alignment between visual features and textual semantics, leading to highly localized and accurate attention maps that closely match ground-truth segmentation. Block-wise, the attention transitions from general domain-level features to fine-grained semantic regions, demonstrating the cumulative impact of each loss component. The final configuration, integrating all loss terms, highlights the synergy between visual and textual modalities, enabling robust cross-domain generalization and precise semantic alignment. These visualizations validate the effectiveness of TCSA-UDA in enhancing attention and segmentation performance through text-driven semantic guidance.

\subsection{Robustness to pseudo-label noise}.
Target-domain prototypes are estimated using pseudo-labels because target annotations are unavailable in the UDA setting. To reduce the influence of noisy pseudo-labels, we adopt three stabilizing strategies. First, target prototype alignment is activated only after a warm-up stage, so early unreliable target predictions are not used for prototype estimation. Second, target pseudo-labels are used to aggregate class-specific features into class-level prototypes, which reduces the influence of isolated erroneous pixel predictions. The prototypes are updated using a momentum rule that allows them to follow the evolving feature distribution during training. Third, the prototype alignment loss is used only as an auxiliary class-level regularizer, together with source supervised segmentation, adversarial alignment, and text-guided semantic regularization. Thus, pseudo-labels are not used as direct pixel-wise supervision for target segmentation, but only to estimate class-level feature centers.
\subsection{Discussion}
The proposed TCSA-UDA differs from existing UDA, prototype alignment, and vision-language medical segmentation methods in several important aspects. Conventional UDA methods mainly reduce domain shift by aligning image appearance, visual features, output entropy maps, or class prototypes. These approaches operate primarily within the visual domain and therefore may struggle when the source and target modalities exhibit substantial appearance differences, such as CT and MRI. In contrast, TCSA-UDA introduces anatomical language semantics as an additional domain-invariant guidance signal. Since anatomical class concepts remain stable across imaging modalities, text embeddings provide a semantic prior that is less affected by modality-specific intensity and texture variations.

The key distinction of VLCoL is that it does not simply align individual visual features with text embeddings. Instead, it aligns the inter-class relationship structure of visual features with the corresponding relationship structure derived from text embeddings. This makes the loss different from direct cosine alignment, contrastive alignment, KL-based matching, or generic visual covariance alignment. Direct image-text alignment treats each class independently, while VLCoL encourages the visual feature space to preserve the semantic geometry among anatomical classes. Similarly, unlike CORAL-style covariance alignment, which matches visual statistics between domains, VLCoL uses language-derived class relations as an external anatomical semantic prior.

The prototype alignment module also has a different role in our framework compared with conventional prototype-based UDA. We do not claim prototype alignment itself as a standalone novelty; rather, it is used as a complementary target-domain adaptation mechanism. VLCoL regularizes the shared encoder with text-derived semantic structure, while prototype alignment reduces residual class-level discrepancies between source and target features. These two components are complementary: VLCoL imposes semantic geometry, and prototype alignment transfers this geometry to the unlabeled target domain.

Compared with existing medical vision-language segmentation methods, TCSA-UDA addresses a different problem setting. Many vision-language segmentation methods focus on supervised or semi-supervised single-domain segmentation, where target labels or paired text-image annotations may be available. In contrast, our method targets unsupervised cross-modality domain adaptation, where target-domain annotations are unavailable. Therefore, text is not used merely as an auxiliary descriptor for segmentation; it is used as a semantic structure prior to guide domain-invariant dense representation learning.

Overall, the uniqueness of TCSA-UDA lies in combining modality-aware anatomical prompting, vision-language relational alignment, adversarial structural adaptation, and prototype-based class-wise alignment in a unified UDA framework. The ablation studies further show that the performance improvement is not due to a single module alone, but to the complementary interaction among text-derived semantic relation learning, visual-language fusion, dynamic feature modulation, and target-domain prototype alignment.

\subsection{Limitations and Future Directions}

Although TCSA-UDA improves cross-modality segmentation performance, some limitations remain. First, the ASD values are not always the best compared with competing methods, indicating that boundary accuracy can still be improved, especially for small or thin structures. Future work will explore boundary-aware losses and shape constraints to reduce surface errors. Second, the current prompts mainly use class, modality, and dataset information. We will investigate LLM-based long-form anatomical descriptors to enrich semantic embeddings and improve cross-domain generalization. Finally, the current framework is mainly based on 2D slice-level adaptation; extending TCSA-UDA to volumetric 3D adaptation is an important future direction to better capture inter-slice anatomical consistency.

\section{Conclusion}
This paper presents TCSA-UDA, a novel unsupervised domain adaptation framework for medical image segmentation that integrates modality-aware text embeddings with visual features to guide semantic alignment through covariance learning and adversarial training. Unlike conventional UDA approaches that rely solely on visual alignment, TCSA-UDA enforces domain invariance at the semantic embedding and feature alignment level, while modality-aware prompts provide context-sensitive guidance without requiring identical formulations. Visual features from heterogeneous modalities are aligned within a shared semantic space, improving cross-domain consistency and preserving class-level discriminability. A semantic prototype alignment module further mitigates residual class-level discrepancies by enforcing class-wise feature consistency between source and target domains. Evaluations on cross-modality cardiac (MRI→CT and CT→MRI), abdominal, and brain tumor segmentation tasks demonstrate that TCSA-UDA improves overall adaptation performance and achieves strong Dice scores across challenging anatomical structures. Boundary accuracy, reflected by ASD, remains an area for further improvement. 

\section*{Acknowledgments}{The authors thank the anonymous reviewers for helpful comments and suggestions. Numerical computations were carried out on the SCIAMA High Performance Compute (HPC) cluster which is supported by the ICG and the University of Portsmouth.}


\end{document}